\documentclass[journal]{IEEEtran}
\usepackage{cite}

\ifCLASSINFOpdf
   \usepackage[pdftex]{graphicx}
\else
\fi
\usepackage{subcaption} 

\usepackage{amsmath}
\usepackage[ruled,vlined]{algorithm2e}

\usepackage{xcolor}

\begin{document}
%
\title{Deep Representation of Imbalanced Spatio-temporal Traffic Flow Data for Traffic Accident Detection}
%
%
%

\author{Pouya~Mehrannia, 
        Shayan~Shirahmad~Gale~Bagi, 
        {Behzad~Moshiri}, 
        {Otman~Adam~Al-Basir}
\thanks{P.Mehrannia and O.A.Basir are with the Department
of Electrical and Computer Engineering, University of Waterloo, Waterloo,
ON, Canada.}
\thanks{S.S.G.Bagi and B.Moshiri are with the Department
of Electrical and Computer Engineering, University of Tehran, Tehran, Iran.}
\thanks{}}

\maketitle

\begin{abstract}
Automatic detection of traffic accidents has a crucial effect on improving transportation, public safety, and path planning. Many lives can be saved by the consequent decrease in the time between when the accidents occur and when rescue teams are dispatched, and much travelling time can be saved by notifying drivers to select alternative routes. This problem is challenging mainly because of the rareness of accidents and spatial heterogeneity of the environment. This paper studies deep representation of loop detector data using Long-Short Term Memory (LSTM) network for automatic detection of freeway accidents. The LSTM-based framework increases class separability in the encoded feature space while reducing the dimension of data. Our experiments on real accident and loop detector data collected from the Twin Cities Metro freeways of Minnesota demonstrate that deep representation of traffic flow data using LSTM network has the potential to detect freeway accidents in less than 18 minutes with a true positive rate of 0.71 and a false positive rate of 0.25 which outperforms other competing methods in the same arrangement.
\end{abstract}


%
\IEEEpeerreviewmaketitle

\section{Introduction}
%
%
%
%


Each year, 1.35 million people die on roadways around the world. Road traffic injuries are estimated to be the eighth leading cause of death globally for all age groups and the leading cause of death for children and young people 5–29 years of age \cite{world2018global}. During the past century, road traffic safety has indicated methods and measures to prevent road accidents. However, an important indicator of survival rates after occurrence of accidents is underestimated. That indicator is the time between when the accident occurs and when emergency medical personnel are dispatched to the scene. Minimizing the time between these two events decreases mortality rates by $6\%$ \cite{white2011wreckwatch}. Researchers have been exploring road infrastructure potentials with the hope to find a reliable approach for real-time sensing of the accident time. A reliable automatic accident detection enables immediate notification to the emergency personnel, and consequently, reduced dispatch time. Moreover, drivers approaching the site can be notified to possibly use alternative routes or at least slow down for safety.

Traffic flow data has been used widely to detect traffic events since the defined metrics on them show much correlation to those events. A considerable portion of the traffic flow data comes from loop detector sensors. These  sensors,  despite  their  noisy  data  and  lack of  high  spatial  coverage,  are  mounted  on  countless  roads. Hence,  developing  models  that  can  yield  good incident detection rates by using these sensors helps governments obtain valuable insights until new technologies start to dominate. Accidents can impact the traffic flow patterns in the upstream and downstream of the accident site. Capturing these patterns in the traffic flow and associating them to the occurrence of accidents is not an easy classification problem. The main objective of this study is to determine the degree to which deep representation of the features obtained from loop detectors can identify crashes automatically after their occurrence.

Another objective is to handle imbalanced spatio-temporal traffic flow data. Traffic accidents are naturally rare events such that the datasets are inevitably imbalanced. The main problem of training a model with imbalanced datasets is that the model cannot provide robust outputs for the minor class since it not exposed to sufficient amounts of data in the minor class to learn the pattern. Moreover, the low number of crash samples can increase the possibility of overfitting in the model.

To fulfil these objectives, We investigate an LSTM-based framework and explore its different settings to increase accident/no-accident class separability in the encoded feature space and to reduce the dimensionality of the traffic flow data. To test our experiments, We use automatic traffic recorder (ATR) data provided by Minnesota department of transportation.  These data are obtained from loop detectors: continuous count devices with loops in the pavement that collect traffic volume. Other useful attributes that are obtained from aggregated ATR data include capacity, density, flow, headway, occupancy, and speed. Since crashes disrupt normal traffic flows, we expect to be able to identify flow disruptions that are attributed to traffic accidents after a certain time past the accidents depending on the nature of them.

The remainder of this paper is organized as a follows. Section \ref{S:2} describes the related work previously done in the field of automatic accident detection. Section \ref{S:3} provides the description of the data and the challenges associated with concurrent using of crash and loop detector data. Section \ref{S:4} presents the modeling approach in detail and provides a short background on the methods incorporated in the model. The model evaluation strategy is also described at the end of this section. Section \ref{S:5} reports and interprets the results. Finally, a discussion on the results and concluding remarks are presented in section \ref{S:6}.

\section{Related Work}
\label{S:2}

Accident detection has been studied in the transportation community for at least 50 years. The California algorithm, developed in 1978, detects the occurrence of accidents when traffic-related variables exceed specific thresholds based on traffic flow theory \cite{payne1978freeway}. Similar algorithms were developed in 70s and 80s by applying the standard normal deviate \cite{dudek1974incident}, Bayesian algorithms \cite{levin1978incident}, and time series models \cite{ahmed1982application} to historical traffic data. The need for a quicker response and handling the interdependencies between input features in the presence of multiple inputs has motivated researchers to resort to more complex frameworks. A wide range of studies targeted real-time accident detection using more sophisticated approaches, including statistical techniques \cite{khan1998statistical,zhang2005towards}, image processing \cite{hoose1992incident,zifeng1997macro}, pattern recognition \cite{rong2013urban,zhang2004new}, and artificial intelligence \cite{abdulhai1999enhancing,adeli2000fuzzy,ccetiner2010neural,jin2001classification,lu2012hybrid}. While many of the proposed methods indicate a reasonable performance using image, social media content, and traffic count data, the viability of these methods in practice using imbalanced, heterogeneous, and real-time data remains uncertain. In many transportation applications, machine learning techniques have proven to handle the presence of high dimensional, heterogeneous, and imbalanced data \cite{lee2018comparison,bortnikov2019accident,parsa2019real}. However, a practical model with a reasonably low response time using the data coming from the already installed traffic infrastructure is still being sought for.

Traffic flow data are being continuously recorded for decades now, hence we normally face big data in this context. Advanced learning methods, like representation learning and deep learning, have shown either promising or much needed for solving the big data problems including traffic accident detection \cite{qiu2016survey}. Among these methods, specific architectures of recurrent neural networks gained increasing attention for their ability to exhibit temporal dynamic behavior and to process variable length sequences of inputs \cite{yadav2020accident}. Unlike classifiers such as support vector machines (SVMs) and logistic regression, deep learning does not seek direct functional relationships between the input features and the output classification results. Instead, it attempts to learn in multiple levels, corresponding to different levels of abstraction \cite{deng2014deep}.

One of the most recent studies employing deep representation for road accident detection is performed by Singh et al. using surveillance videos \cite{singh2018deep}. Their proposed framework automatically learns feature representation from the spatiotemporal volumes of raw pixel intensity which outperformes traditional hand-crafted features based approaches. Another study by Zhang et al. \cite{zhang2018deep} employed deep belief network (DBN) and long short-term memory (LSTM) in detecting the traffic accident from social media data. Their results show that their method outperforms support vector machines (SVMs) and supervised Latent Dirichlet allocation (sLDA) tested on 3 million tweet contents in two metropolitan areas in a 1-year time window. Hunag et al. \cite{huang2020intelligent} used an integrated two-stream convolutional networks architecture to detect near-accident events of road users in traffic video data. Their real-time deep-learning-based method comprises a spatial stream network and a temporal stream network, and shows competitive qualitative and quantitative performance at high frame rates. Ghosh et al. \cite{8816881} also used videos from a CCTV camera installed on a highway to train a  Convolutional Neural Networks (CNN) based model to classify individual frames to crash or non-crash events.

Traffic data collecting technologies are evaluated mainly based on reliability, performance under various environmental conditions, the accuracy of obtained data from detectors, and performance in real time \cite{nikolaev2017analysis}. Video image processing has the advantage of providing images of movement in real-time and no traffic interruption for installation and repair. However, video contents, just like social media contents, are expensive to access and process and the performance of models working based on them in many cases have shown inconsistent under various environmental and traffic circumstances. The usage of smartphones and GPS-based applications are also popular for their good accuracy and real-time performance, but smart phones have privacy complications, and due to false alarm filter, it may not detect all accidents \cite{amin2012accident}. In-vehicle emerging technologies appearing in, for example, eCall , Onstar, and Ford-Sync \cite{smruthi2020intelligent, lupinska2020car} can automatically make an emergency call after an accident. These services require cross-border interoperability, because vehicles cross national borders and because of the global and regional nature of the automotive market. Also, they must be physically robust to survive an accident, and since the expected lifespan of vehicles is longer than phones’, backward compatibility with previous generations of cellular networks is required \cite{oorni2017vehicle}. In this study we use loop detectors as they are installed in abundance on roadways and are not sensitive to environmental and traffic factors such as rain, snow, and traffic jams. Despite collecting noisy data and lack of high spatial coverage, loop detectors, their installation, and their maintenance are relatively low cost and provide a large knowledgebase with relatively good performance \cite{nikolaev2017analysis}, which motivate us to find efficient approaches that can work based on loop detector data.

There are other challenges when dealing with accident data. Road accident datasets are highly imbalanced which yields biases down the road when the classification model is being trained. The number of non-accident samples is much larger than the number of accident cases in real accident datasets. The proportion of accident and non-accident records in the dataset needs to be meticulously assigned \cite{katrakazas2016real}, or proper oversampling and undersampling methods should be considered to deal with imbalanced data \cite{ozbayoglu2016real,mujalli2016bayes}. Another powerful method, synthetic minority over-sampling technique (SMOTE) \cite{chawla2002smote}, has shown to be efficient when only a few samples of one class are available, which is the case in accident datasets. Synthesizing minority class samples results in larger and less specific decision regions which resolves the overfitting issue of oversampling and the datapoint loss issue of undersampling. Generally, it has been observed that SMOTE is more robust, especially when dealing with noisy, large, and sparse datasets \cite{kaur2018comparing,vanhoeyveld2018imbalanced}, yet no satisfying feature representation method has been proposed in combination with these resampling techniques to process traffic count sensor data. Deep representation of traffic flow data incorporating networks, like LSTM or CNN, enables accident detection models to learn the spatio-temporal patterns in data and their dependencies to accident events \cite{wang2020deep}. A recent successful application of deep representation of data using the LSTM network is for short-term traffic forecast by Zhao et al. \cite{zhao2017lstm}.  Finding the proper combination of deep representation and resampling methods along with the proper network's structure and settings is a crucial task to overcome these challenges. We have investigated and compared these combinations and are proposing the one with the best outcome in section~\ref{S:4}.

\section{Data Preparation}
\label{S:3}

The data for this study is provided by the Minnesota Department of Transportation (MnDOT) which includes the traffic count data and the accident records data \cite{MnDoT} for a duration of 10 years from 2009 to 2019.

\subsection{Traffic Count Data}
\label{S:31}

The loop detector data are continuously collected by the regional transportation management center (RTMC), a division of MnDOT, at a 30-second interval from over 4,500 loop detectors located around the Twin Cities Metro freeways, seven days a week and all year round. All detectors produce vehicle volume data; however, some automatic traffic recorder stations collect additional types of data depending upon their equipment and sensors. In this study, we use 6 features extracted from MnDOT detector tools, namely: capacity, density, flow, headway, occupancy, and speed described as follows:

\begin{itemize}
    \item [--] Occupancy is the percentage of time a detector's field is occupied by a vehicle.
    \item [--]	Flow is the number of vehicles that pass through a detector per hour $(Volume \times Samples \; per \; Hour)$. 
    \item [--] Headway is the number of seconds between each vehicle $(Seconds \; per \; Hour$ / $ Flow)$
    \item [--] Density is the number of vehicles per mile
    \item [--] Speed is the average speed of the vehicles that pass in a sampling period $(Flow / Density)$
    \item [--] Capacity is the amount of flow that can be added (or subtracted) to a roadway before reaching high congestion (or low traffic)
\end{itemize}

\begin{figure}[t]
\centering\includegraphics[width=1\linewidth]{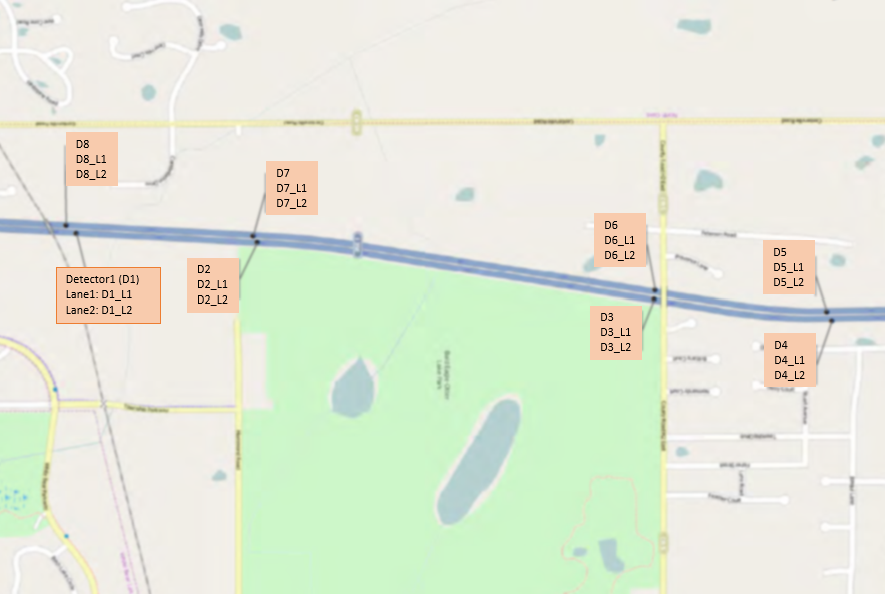}
\caption{A bidirectional traffic site containing 16 sensors.}
\label{fig:figure1}
\end{figure}

We perform our experiments on the sections of freeways relatively far from ramps. More complex freeway structures and roads with intersections are left for future work. Figure~\ref{fig:figure1} shows an example site used for collecting detector data. This particular site is equipped with 16 sensors, each of which provides the features of interest for the lane that it is installed in. These sensors aggregate data in a preset interval of 30 seconds, but the data can be extracted in intervals of 0.5, 1, 1.5, 2, 2.5, 3, 4, and 5 minutes. The distances between detectors along the direction of the traffic flow are not the same. However, the average distance in the experiment sites is approximately 0.8 mile which is low enough to capture the fluctuations in the traffic flow by at least one sensor in the proximity not long after occurrence of an accident. The severity of an accident is also a factor that can affect the time to detection of the accident. We will discuss the time to detection in the results section and also the type and configuration  of the accidents that are difficult to detect by the proposed systems.

\subsection{Road Accident Data}
\label{S:32}

The accident records for the experiment sites were used to test the performance of the proposed method in this study. This accident dataset has the records of all accidents in the experiment sites in the past 10 years. The records include the date and time stamps of the accidents which are used to evaluate the accuracy of accident detection. A quick analysis of the accident time stamps in the dataset revealed that not all of them are accurate. Our understanding is that the accident time stamps are recorded based on the evidence found in the crash site or reported by witnesses. Sometimes, there are not enough evidence or witnesses available to capture an accurate accident time stamp. A comparative analysis of the accident data against the sensors data showed that the time stamps can have a shift of tens of minutes in some instances. Unfortunately, this inaccuracy can affect the evaluation of accident detection model, especially when minimizing the time to detection criterion is of interest. In order to decrease the effect of inaccurate time stamps, we added a degree of flexibility (DoF) parameter that helps to better reflect on the correctly identified accidents and the false alarms. We explain about how we adopted DoF in the methodology chapter in more details.

\section{Methodology}
\label{S:4}

Traffic accident detection is a binary classification problem in which there are two classes: crash and non-crash. There are two remarkable points about traffic accidents. First, traffic accidents are naturally rare events such that the datasets are inevitably imbalanced. The main problem of training a network with imbalanced dataset is that the network cannot provide optimal results for the minor class since it does not learn sufficient amounts of patterns in the minor class. To deal with this problem, we increase the weight of the samples in the minor class when training the network, and resample the dataset to generate more instances of the minor class.

\begin{figure}[t]
 \begin{subfigure}{.5\textwidth}
    \centering
    \includegraphics[width=0.8\linewidth]{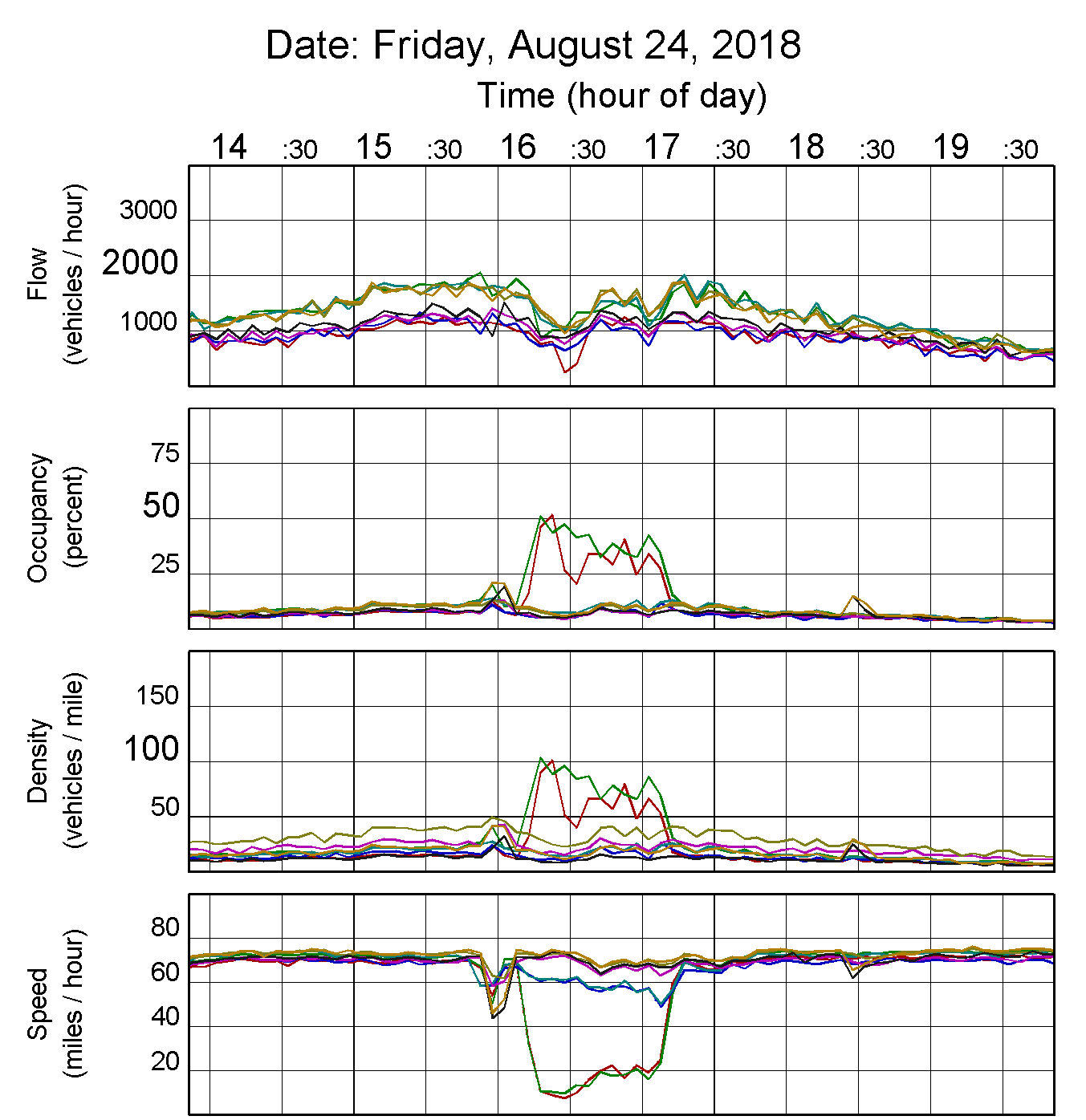}
    \caption{Crash}
    \label{fig:pattern_accident}
\end{subfigure}%
    
\begin{subfigure}{.5\textwidth}
    \centering
    \includegraphics[width=0.8\linewidth]{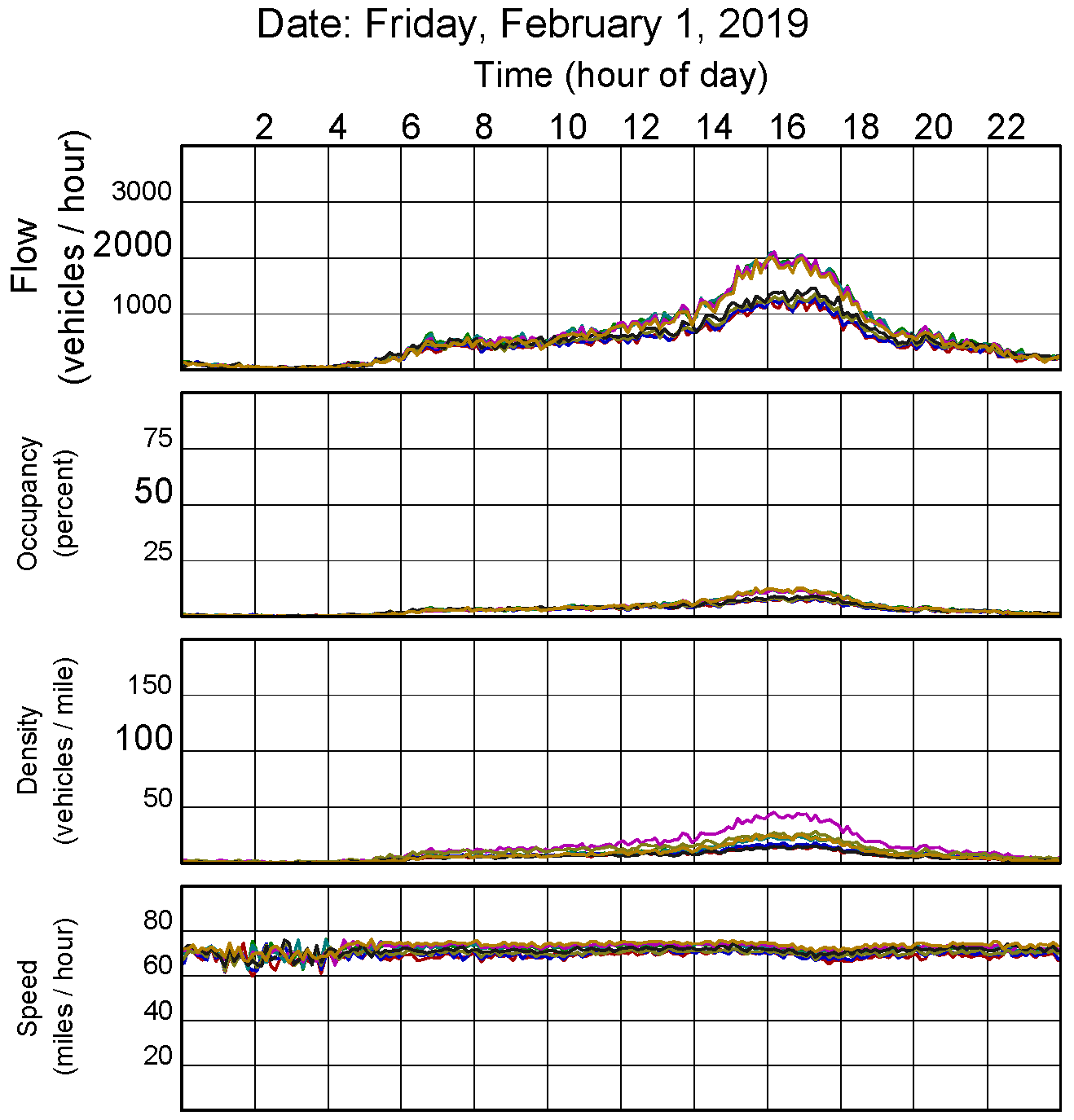}
    \caption{Traffic jam}
    \label{fig:pattern_trafficjam}
\end{subfigure}%

\caption{Patterns of sensor reading features before and after a crash and a rush-hour traffic jam that happened at 4:00 pm. Each color represents a particular sensor in the vicinity of the crash location.}
\label{fig:pattern}
\end{figure}

Secondly, when a crash occurs, the density and speed of cars are affected in both directions of the crash site \cite{ golob2004method}. The patterns that are generated for the speed and density are sometimes modeled by simple distributions, e.g. normal or lognormal distribution, and sometimes by more complex models depending on the importance of the details in the simulation \cite{mehrannia2018dempster}. If there are traffic count sensors installed in the region where an accident has happened, these sensors would be able to capture the patterns if their sampling rate are high enough. Figure~\ref{fig:pattern} shows the speed, density, occupancy, and flow patterns captured by 8 adjacent sensors after occurrence of a crash (Figure~\ref{fig:pattern_accident}) and a rush-hour traffic jam (Figure~\ref{fig:pattern_trafficjam}) on a freeway.
 
The temporal and spatial patterns of a crash can be clearly seen in Figure~\ref{fig:pattern_accident}. This figure illustrates different features of the sensors in the crash site. The sensors, represented by different colors, are spatially spaced, and their features shows that the average speed of vehicles in the crash area has been reduced drastically at the crash time. Furthermore, the traffic accident affects the sensors data in a certain period of time depending on the nature of the accident and the crash site characteristics. Traffic data are spatio-temporal; meaning that traffic parameters such as flow vary both in time and space. Therefore, it would be reasonable to consider time series of multiple adjacent sensors data to detect the occurrence of a traffic accident. There is one major difference between a traffic jam and an accident and that is their spatial patterns. In a traffic jam, caused by the rush hours of the day for example, all lanes of the highway would be crowded, hence, the traffic parameters of all loop detectors would fluctuate simultaneously, as shown in Figure~\ref{fig:pattern_trafficjam}. In contrast, when an accident occurs, it has exclusive effect on each of the loop detectors' readings. For example, in Figure~\ref{fig:pattern_accident}, occupancy, density, and speed measured by two of the loop detectors change significantly after the crash while they change only slightly in the other two.

 \begin{figure*}[!t]
\centering\includegraphics[width=\textwidth]{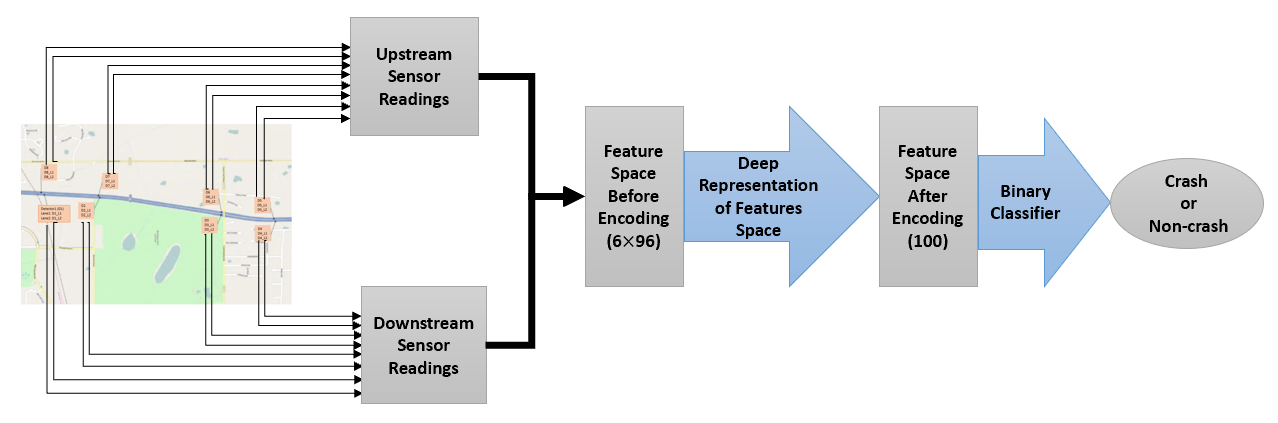}
\caption{An overview of the main steps for automatic detection of freeway accidents. }
\label{fig:genov}
\end{figure*}

Figure~\ref{fig:genov} shows an overview of the main steps to detect crashes by looking at 16 adjacent sensors data (8 in each direction) in consecutive time intervals. This classification problem can be divided into two phases: encoding phase and classification phase. In the encoding phase, a neural network learns which features are most significant for classification phase. This phase also eliminates correlations between the features in order to reduce the dimension of data and obtain a more efficient feature space. By creating a feature space with better separability, this step is crucial in applications with complex data. We will show that the proposed network is able to automatically learn the efficient features by comparing the feature spaces before and after encoding. Since LSTM has high capability in handling time series data, we incorporate it in our study for the encoding phase. It is evident from Figure~\ref{fig:pattern} that accidents create spatio-temporal patterns in traffic parameters such that loop detectors data are affected both through time and space. Furthermore, these patterns can be seen as sequences i.e. when accidents occur, flow increases over time then remains constant and then reduces over time. Any network has to learn these patterns in order to detect the accidents. LSTMs are capable of learning such dependencies by design and are used broadly for processing sequences.

Other memory neural networks such as simple recurrent neural networks (RNNs) are not as suitable for accident detection since they do not have long-term dependencies to previous samples. This is due to the gradient vanishing problem that appears in back propagation through time. In our study, we also encode the feature space using Convolutional Neural Network (CNN) and other conventional methods to compare the proposed network’s capability in encoding feature space with theirs. A brief explanation of the resampling and encoding methods are provided in the following sections.

In the classification phase, the network learns to correctly detect a traffic accident using the features provided by the encoding phase. This phase can be implemented using various artificial neural networks, such as multi-layer perceptron (MLP), probabilistic neural network (PNN), and radial basis function (RBF) network. In our study, we use the MLP for the classification task for several reasons; Unlike RBF networks, MLPs are easier to train and can also classify non-convex feature spaces. Also, unlike PNNs which are feed forward neural networks, MLPs are trained using backpropagation algorithm, hence can be used in complex problems dealing with high dimensional data. Our proposed algorithm for accident detection is summarized in Algorithm~\ref{Alg:accdet}.

\begin{algorithm}[h]
\DontPrintSemicolon
\KwIn{Sensors raw data (Flow, Density, Speed, Capacity, headway, Occupancy)}
\KwOut{Class of sample (crash or non-crash)}
\;
\nl  Discarding irrelevant crashes and cleansing data, \;
\nl Data partitioning and splitting test and train data, \;
\nl Normalizing dataset using Z-normalization (zero mean and unit variance), \;
\nl Resampling training data, \;
\nl Time series generation, \;
\nl Encoding feature space using LSTM\;
\nl Classifying the encoded feature space and determine crash probability, \;
\nl Compare the crash probability with a threshold to decide whether an accident has occurred or not.

 \caption{Algorithm of accident detection}
 \label{Alg:accdet}
\end{algorithm}

First, we discard the crashes that have not occurred on the target highway, e.g., crashes on the roads that are in close vicinity or intersect the highway. This step is needed since the crash dataset used for training is populated based on GPS location. In practice, for the run of the trained model, this step becomes unnecessary since the input data is pulled from the loop detector data only. We also cleanse the data of malfunctioning loop detectors according to the acceptable ranges stated in the documents of MnDOT Regional Traffic Management Center (RTMC). The dataset contains accidents with different severity levels which can create different spatio-temporal patterns in loop detectors data. To ensure that the proposed model can achieve the desired performance in detecting all types of accidents, we use k-fold cross validation to split the dataset. By doing so, all types of accidents will be contained in test data. After normalizing data, the data is resampled to balance the number of samples in each class. In the sixth step, we encode the spatio-temporal features using LSTM to obtain a more efficient feature space. This step is crucial because it reduces the dimension of data and increases the class separability which contributes to higher detection rate. Finally, we determine the crash probability by using a sigmoid function in the output layer of the network. If the crash probability of a sample is higher than 0.5, it is decided that an accident has occurred, otherwise the sample is labeled non-crash.

\subsection{Resampling Methods}
\label{SS:42}

Resampling methods are in one of the categories of under-sampling methods, over-sampling methods, or a combination of both. Under-sampling methods balance the dataset by reducing the size of major class’s samples while over-sampling methods balance the dataset by increasing the size of minor class’s samples \cite{yuan2019real}. Some of these methods are explained below.

\begin{itemize}
    \item[-] Edited nearest neighbor (ENN): In this method, the points whose class label is different from majority of its $k$ nearest neighbors are removed \cite{more2016survey}.
    
    \item[-] Tomek Link Removal: In this method, the points that are in vicinity of each other and belong to different classes which is called a Tomek link are removed from dataset. As a substitute, one can only remove points belonging to the major class from Tomek link \cite{more2016survey}.
    
    \item[-] Synthetic Minority Oversampling Technique (SMOTE): In this method, first, $k$ nearest neighbors of point $p$ which belong to the minor class, are computed. To generate a synthetic point for each point $p$, a minor class point $p_z$  is randomly chosen from $k$ nearest neighbors of point $p$ and then the new synthetic point is calculated as follows \cite{chawla2002smote}:
    
    \begin{equation}
    \label{eq:e1}
    x_{new} = p + \rho(p_z-p)
    \end{equation}
    where $\rho$ is a random number in the range [0,1]. There are some variations of SMOTE such as Borderline SMOTE, SVM-SMOTE, and Adaptive Synthetic Sampling (ADASYN) which are discussed in details in \cite{more2016survey}, \cite{nguyen2011borderline}, and \cite{he2008adasyn}.

    \item[-] Combinational Methods:  These methods combine different resampling methods to improve efficiency in certain scenarios. Examples of  Combinational Methods are SMOTE-ENN  \cite{more2016survey} and SMOTE-Tomek link removal \cite{more2016survey}.

\end{itemize}

\subsection{Deep Representation of Features}
\label{SS:43}

\begin{figure}[t]
\centering\includegraphics[width=0.95\linewidth]{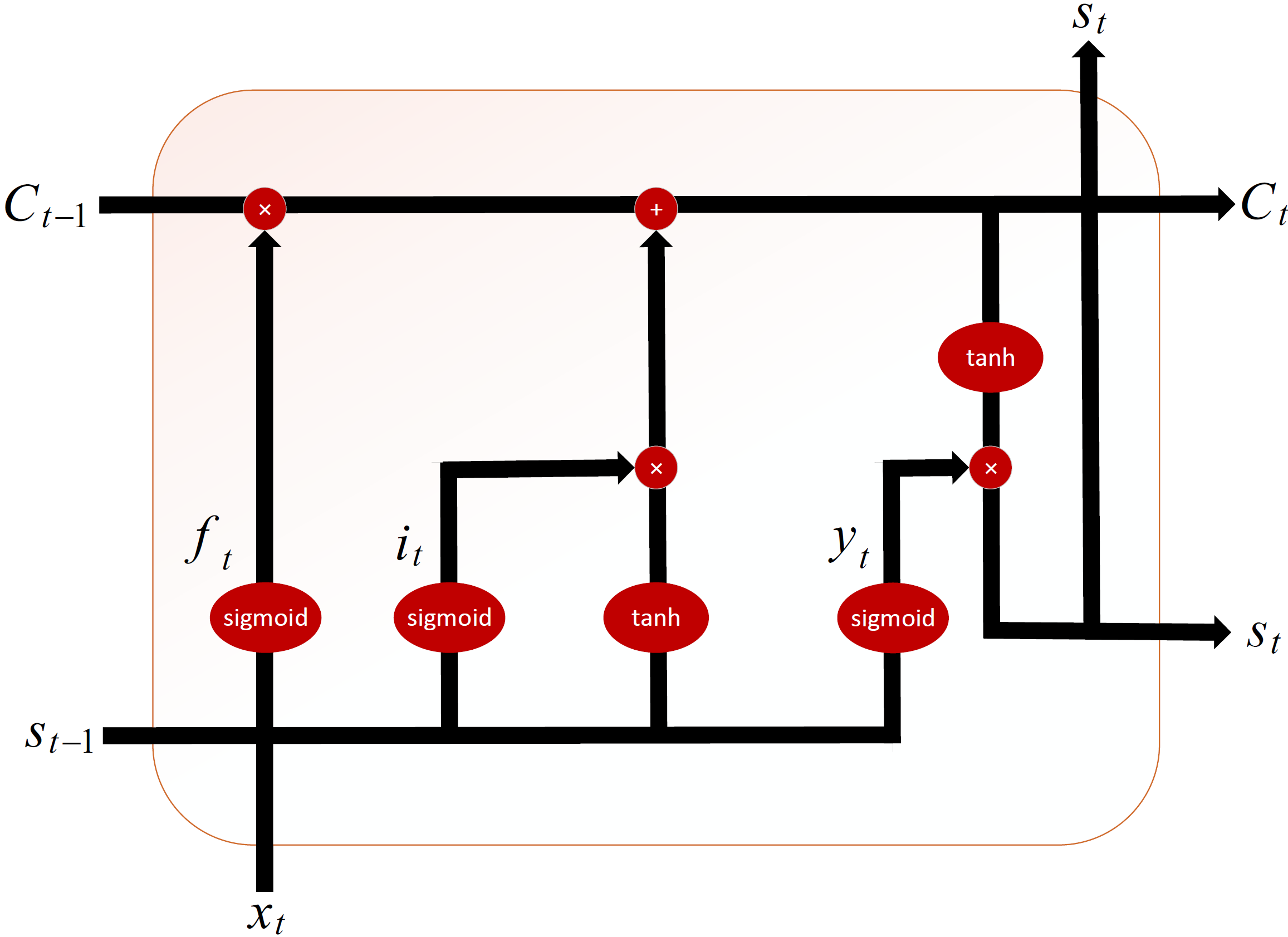}
\caption{LSTM cell}
\label{fig:LSTM}
\end{figure}

In this study, we use LSTM to map the high dimensional features of loop detector data to a new optimized feature space with reduced features. LSTM consists of two main units, namely, gates and cell state. Input and output gates grant control over inputs and outputs of the cell, protecting it from any irrelevant perturbations \cite{hochreiter1997long}. On the other hand, cell state allows constant error flow through units to avoid gradient vanishing problem. Cell state and gates, together, form an LSTM cell as shown in Figure~\ref{fig:LSTM}.

There are three main operations in gates of LSTM cells. In the forget and the input gates, it is decided that what information should be deleted or stored in the cell state, respectively. Finally, in the output gate, it is decided that what the memory cell is going to output. These operations are summarized in the equations below \cite{hochreiter1997long}:

\begin{equation}
\label{eq:e8}
i_t = \sigma(W_{ix} \times x_t + W_{is} \times s_{t-1} + b_i ) 
\end{equation}

\begin{equation}
\label{eq:e9}
f_t = \sigma(W_{fx} \times x_t + W_{fs} \times s_{t-1} + b_f )
\end{equation}

\begin{equation}
\label{eq:e10}
c_t = f_t \odot c_{t-1} + i_t \odot \tanh (W_{cx} \times x_t + W_{cs} \times s_{t-1} + b_c)
\end{equation}

\begin{equation}
\label{eq:e11}
o_t = \sigma(W_{ox} \times x_t + W_{os} \times s_{t-1} + W_{oc} \times c_t + b_o )
\end{equation}

\begin{equation}
\label{eq:e12}
s_t = o_t \odot \tanh{c_t}
\end{equation}
where $i_t$, $f_t$, $o_t$, $c_t$, $s_t$ denote for input gate, forget gate, output gate, memory cell, and internal state at time step $t$, respectively. $\sigma$ is the logistic sigmoid function and $\odot$ indicates Hadamard product.

\subsection{Network Architecture}
\label{SS:44}

\begin{figure}[t]
\centering\includegraphics[width=1\linewidth]{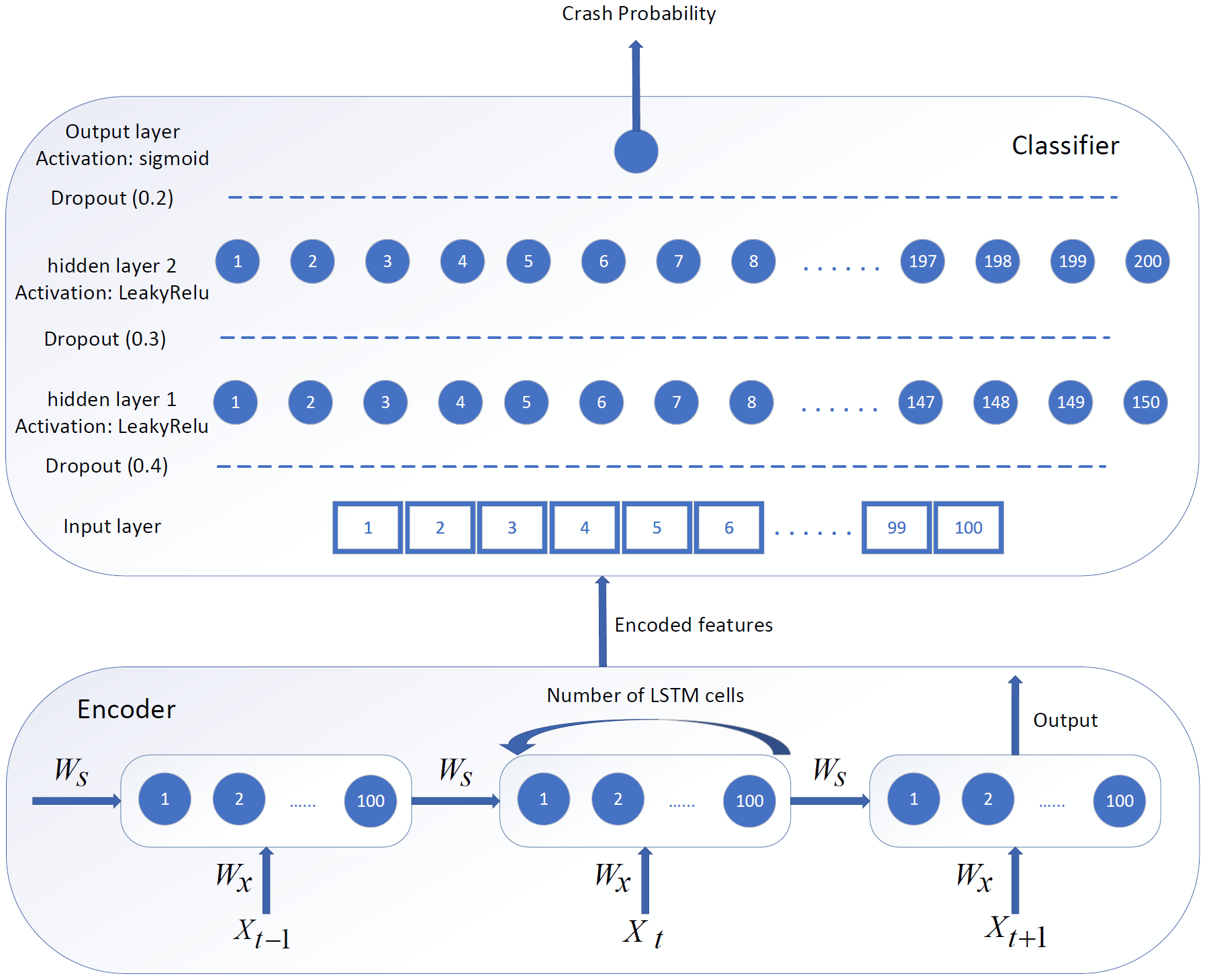}
\caption{proposed network architecture}
\label{fig:arch}
\end{figure}

Our proposed network includes an encoder and a classifier. As explained previously, we use MLP as the classifier and LSTM as encoder. It should be noted that the proposed network is similar to the well-known LSTM and CNN, however, we divide our network into encoder and classifier part to emphasize that the network performs two different tasks: feature learning and classification. An illustration of proposed network is given in Figure~\ref{fig:arch}.  

Figure~\ref{fig:arch} shows the layers used in our proposed network with their corresponding size and activation functions. There are three types of layers in the network: LSTM layer, dropout layers, and fully connected layers. The main motivation for using dropout layers is to prevent model from overfitting. There are few reasons for choosing Leaky ReLU as the activation function of fully connected layers. First, ReLU and Leaky ReLU does not bound the outputs of the layers, hence, they can reduce the effect of gradient vanishing problem. Secondly, the dataset is normalized using the Z-normalization method and it contains negative values. Consequently, when using ReLU, some neurons may be inactive no matter what input is being supplied, which prevents the gradient flow. Therefore, it is better to use Leaky ReLU instead of ReLU as the activation function. As shown in Figure~\ref{fig:arch}, first, input feature space is encoded using LSTM and then encoded features are passed on to the fully connected MLP layers. It is worthwhile noting that since our objective is to detect crashes and not predicting them, the time series data should be passed along to the encoder in reverse chronological order, e.g., first input to the encoder would be the last time slice. In other words, we detect accidents after they occur, hence, we present the data to the network backwards in time. The classifier has two hidden layers, because MLP with two hidden layers is more effective than MLP with one hidden layer in classifying non-convex feature spaces. The dimension of each layer in the classifier is greater than the one in the previous layer except for the output layer which only has one neuron, because there are only two classes. The size of each layer has been fine-tuned empirically, however, according to the Cover’s theorem, linear separability of the feature space increases relatively to the space dimension.

\subsection{Training the Model}
\label{SS:45}

The proposed network is trained using backpropagation algorithm. To further deal with the imbalanced dataset and prevent overfitting, we use weighted binary cross-entropy loss function as expressed in Equation~\ref{eq:e13}:

\begin{equation}
\label{eq:e13}
L(y,\hat{y}) = -\frac{1}{N} \sum_{i=1}^{N} (w_1*y_i*\log (\hat{y_i})+w_2*(1-y_i)*\log(1-\hat{y_i}))
\end{equation}
where $y_i$ is the true label of the $i^{th}$ sample, $\hat{y_i}$ is the predicted probability of crash, N is number of samples, and $w_1$ and $w_2$ are class weights. By assigning high weight to the minor class which includes crash samples, the network puts more effort to correctly classify the samples of this class. The penalty of incorrectly predicting the label of these samples would be high when training the network using backpropagation. Consequently, even when network can detect non-crash samples correctly, the loss function’s value will not diminish which is desirable for training the network.

\subsection{Model Evaluation}
\label{SS:46}

In the crash detection problem, it is possible to achieve high accuracy while true positive rate remains low due to imbalanced classes. However, it is essential to achieve high true positive and low false positive rates. Therefore, we will mainly focus on sensitivity and specificity when evaluating the models. As discussed in Section \ref{S:3}, in order to compensate for any errors in recording of crash events, we consider a Degree of Flexibility (DoF) in calculation of confusion matrix. By taking DoF into account, True Positives (TP), False Negatives (FN), True Negatives (TN), and False Positives (FP) are calculated as follows:

\scriptsize
\
\begin{equation}
\label{eq:e14}
\hspace{-0.85em}
  \begin{cases}
      TP_j=TP_{j-1}+1 & \text{if \{$t(j)=1$ and $P > 0$\}}\\
      FN_j=FN_{j-1}+1 & \text{if \{$t(j)=1$ and $P = 0$\}}\\
      TN_j=TN_{j-1}+1 & \text{if \{$t(j)=0$ and $p(j)=0$ or $t(j)=1$ and $T > 0$\}}\\
      FP_j=FP_{j-1}+1 & \text{if \{$t(j)=0$ and $p(j)=1$ and $T = 0$\}}
    \end{cases}       
\end{equation}
\normalsize
where $$P=\sum\limits_{n=j-DoF}^{j+DoF} p(n),$$ $$T=\sum\limits_{n=j-DoF}^{j+DoF} t(n),$$ and $t(j)$ and $p(j)$ denote for true label and predicted label of $j^{th}$ sample, respectively.
Sensitivity determines network’s ability to correctly predict crash samples:

\begin{equation}
\label{eq:e16}
\text{Sensitivity} = \frac{TP}{TP+FN}
\end{equation}

Specificity determines network’s ability to correctly predict non-crash samples:

\begin{equation}
\label{eq:e17}
\text{Specificity} = \frac{TN}{TN+FP}
\end{equation}

\section{Experimental Results and Discussion}
\label{S:5}

In this section, we evaluate the LSTM-based accident detection method along with other competing methods and report their performance from various aspects. For the encoder implementation phase, the networks are trained using different optimization algorithms, resampling methods, time series length, and data aggregation interval. Furthermore, an analysis of the LSTM-based network’s capability in encoding feature space and a comparison between detection rate of this network and other classifiers including CNN, PNN, AdaBoost, and GradientBoost is provided in this section. A 5-fold cross validation procedure is applied to the data gathered from the sensors of the site shown in Figure~\ref{fig:figure1}, such that the data is divided into 5 partitions randomly, and each time, 4 of the partitions are used for training and the other for testing. In the testing process, a confusion matrix is generated to show the number of correctly identified accidents and false alarms. The average accuracy, sensitivity, and specificity of the 5 folds are reported to have unbiased evaluation criteria. 

\subsection{Model Performance with Different Optimization Algorithms and Resampling Methods}
\label{SS:51}

 \begin{figure}[t]
\centering\includegraphics[width=0.95\linewidth]{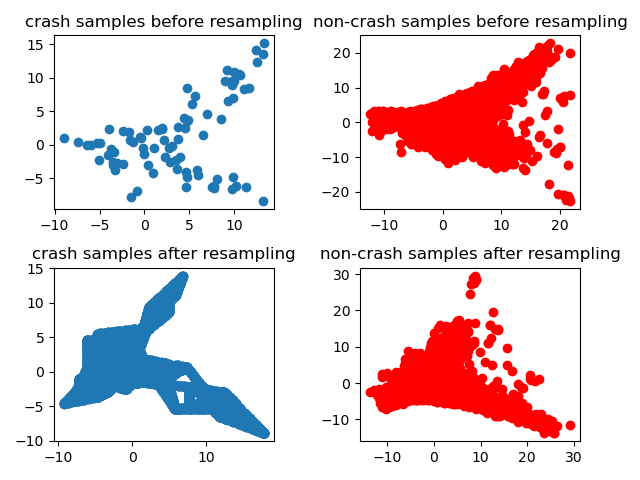}
\caption{training data before and after resampling}
\label{fig:resam}
\end{figure}

In this section, we only focus on the performance of resampling methods and optimization algorithms. The impact of other significant factors such as time series length are restricted by tuning them in their best possible arrangement. The first step after preprocessing the accident dataset is the resampling of the training data to deal with its imbalanced nature. To do so, we should determine which resampling method is more suitable to the case study. To see the impact of resampling on the training data, a 2D visualization of the feature space before and after the resampling is represented in Figure~\ref{fig:resam} using principal component analysis (PCA).

 \begin{figure}[t]
\centering\includegraphics[width=0.9\linewidth]{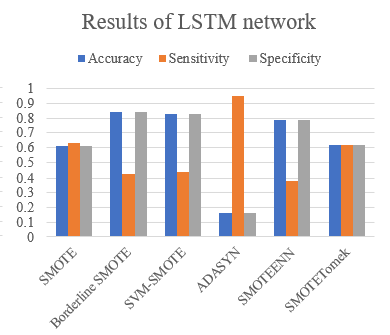}

\centering\includegraphics[width=0.95\linewidth]{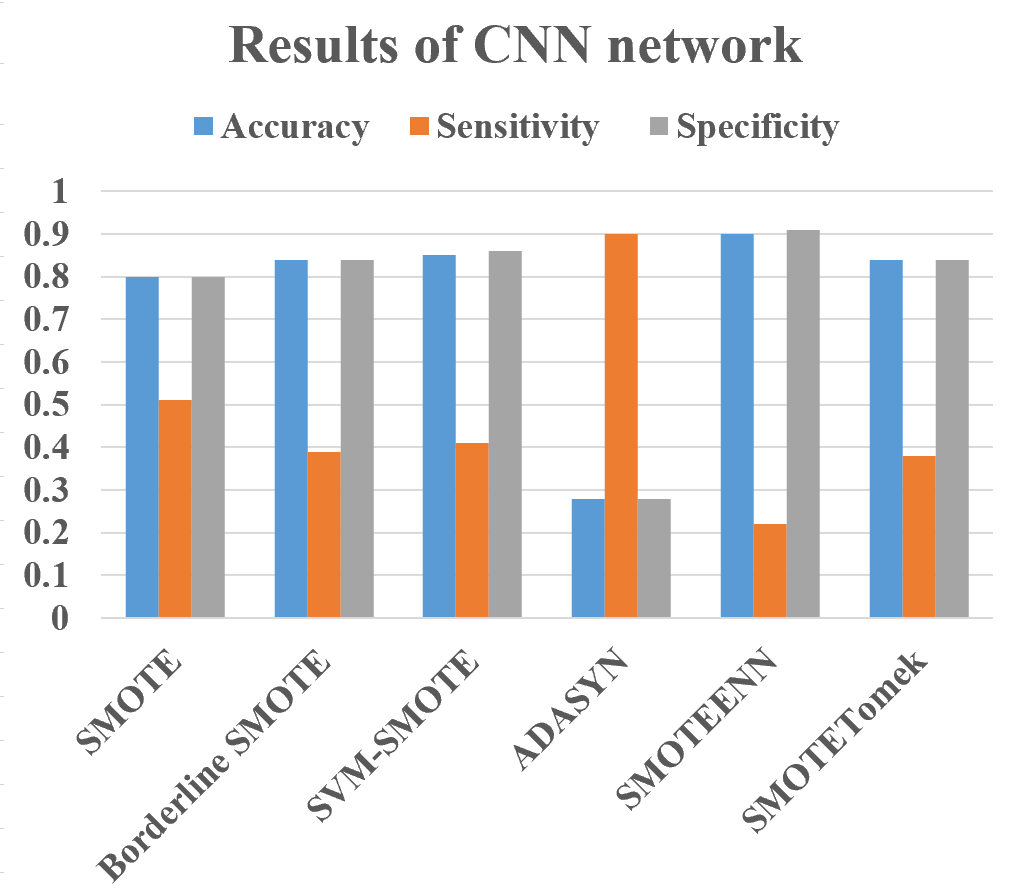}
\caption{results of the LSTM- and CNN-based models trained using different resampling Techniques}
\label{fig:resamres}
\end{figure}

As explained in Section~\ref{S:4}, resampling methods balance the dataset by increasing the samples of minor class and/or reducing the samples of major class. This is why some crash samples have been generated such as the point with coordinates (-10, -5) and some non-crash samples have disappeared such as the point with coordinates (20, -20). It should be noted there are also some outliers generated by resampling methods such as the point with coordinates (10, 30) in non-crash samples. It should be determined that which resampling method works best in our problem. To do that, a comparison between the results of models trained with different resampled training data is performed. The results of the top two models (LSTM and CNN) are shown in Figure~\ref{fig:resamres}. It should be noted that resampling methods generate artificial data, hence, they might not be realistic. Nonetheless, they balance the dataset and eliminate the bias toward major class. In order to provide reliable results, we have left the test data untouched in all experiments.

As illustrated in Figure~\ref{fig:resamres}, ADASYN and SMOTE have better sensitivity and specificity compared to the other methods. Although, ADASYN outperforms all other methods in terms of sensitivity, its specificity is too low which increases the number of false alarms drastically. SMOTE, however, has a good balance of accuracy, sensitivity, and specificity. Therefore, SMOTE will be used to resample the training data in the next parts. It is observed that high accuracy does not guarantee high sensitivity due to low number of crash samples. Sensitivity is a good metric for measuring network’s ability in classifying accidents, however, it cannot give insights about false alarm rate of the network. Therefore, specificity is required to measure network’s ability in classifying non-crash samples.  So, the next evaluations will be based on sensitivity and specificity only.

The next step is to decide which optimizer provides the best outcome. To do so, we evaluated the performance of adaptive learning methods and non-adaptive learning methods in our accident detection case study. Stochastic Gradient Descent (SGD) with momentum is used here to represent non-adaptive learning methods. The reason we use momentum is to speed up the learning process and escape from the suboptimal local minima of the loss function. From now on, we will refer to SGD with momentum simply as SGD. Adaptive Moment Estimation (ADAM) \cite{kingma2014adam} performs well in practice and compares favorably to other adaptive learning methods such as Adagrad \cite{duchi2011adaptive}, Adadelta \cite{zeiler2012adadelta}, and RMSprop, hence, it is chosen to represent the adaptive learning methods. The proposed network is trained using ADAM and SGD optimizers with a learning rate of 0.00001, momentum of 0.8, batch size of 32, and class weight of 1:100 for 70 epochs. The values of these hyperparameters have been tuned empirically and will be used in all proceeding experiments.

\begin{table}[t]
\centering
\caption{Comparison of optimization algorithms for LSTM}
\vspace{-0.5\baselineskip}
\begin{center}
\begin{tabular}{|c| c| c| c| c| c| c| c|}
\hline
 & Accuracy & Sensitivity & Specificity & Training Time (s) \\
 \hline

 SGD  & 0.61  & 0.63  & 0.61 & 569 \\
 \hline
 ADAM  & 0.73  & 0.37  & 0.73 & 581 \\ 
\hline

\end{tabular}
\end{center}
\label{table:1}
\end{table}

The low number of crash samples can increase the possibility of overfitting in the models. According to Table~\ref{table:1}, sensitivity of the LSTM-based model reduces drastically when using ADAM optimizer. This happens because SGD has better generalization capability compared to ADAM. SGD is also computationally more efficient than ADAM based on the training times reported in Table~\ref{table:1}. In the next parts, we use SGD as the selected optimizer and report the results of evaluation for the other criteria.

\subsection{Model Performance with Different Data Aggregation Intervals and Time Series Length}
\label{SS:52}

 \begin{figure*}[!t]
\centering\includegraphics[width=0.6\linewidth]{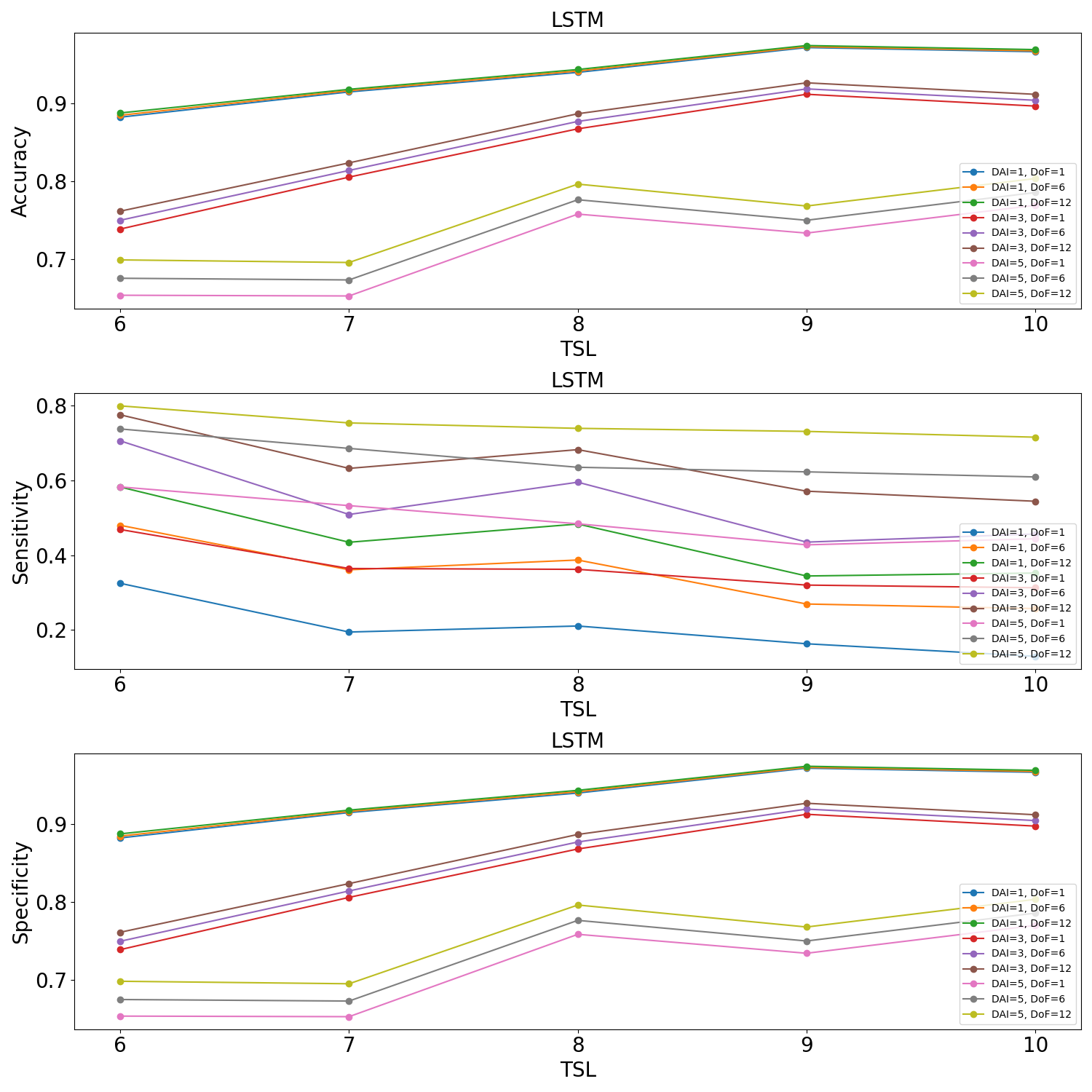}
\caption{LSTM performance for various TSL, DAI, and DoF}
\label{fig:result-LSTM}
\end{figure*}

Time to detection (TTD) is an important aspect of an accident detection model. TTD is the time needed by the model to be able to detect an accident after it occurs. A small TTD value is essential along with high sensitivity and specificity in accident detection models. The key parameters affecting TTD are data aggregation interval (DAI), time series length (TSL), and degree of flexibility (DoF). DAI is a tunable parameter in the data extraction tool. It has an impact on the smoothness of fluctuations in traffic data and together with TSL, they determine how many temporal samples are required to detect an accident. On the other hand, DoF accounts for model’s delay in detecting the time of accident which can be imposed by inaccurate crash records in the dataset.

According to Figure~\ref{fig:result-LSTM}, specificity and sensitivity of the LSTM-based model increase with DoF. This is trivial since more delayed detections are accepted as DoF increases. Both models have achieved high specificity, so the first priority in determining the optimal TSL and DAI values is sensitivity. Having a look at the plots with DoFs of same value (i.e. DoF =1 or 6, or 12), it can be concluded that models trained with higher DAI have higher sensitivity. Therefore, there must be a trade-off between TTD and sensitivity of the models. Furthermore, sensitivity values are non-monotonic decreasing functions of TSL, hence the optimal time series length would be 6. Therefore, we will compare models Receiver Operating Characteristic (ROC) at TSL=6. Maximum sensitivity and specificity achieved by the LSTM model using optimal parameters are 0.8 and 0.7, respectively.

\subsection{Capability in Encoding Feature Space}
\label{SS:53}

As explained before, the main motivation for using LSTM network in our study was to encode the spatio-temporal features and obtain a more efficient feature space. In this section, we are going to demonstrate this network’s capability in encoding the feature space by comparing the separability of classes in 3 feature spaces: original feature space (before encoding), encoded feature space with LSTM, and encoded feature space with CNN. The reason we choose CNN for comparison here is that CNN can also encode the spatio-temporal features with 1D convolutional layers and can handle time series data.

 \begin{figure}[t]
\centering\includegraphics[width=0.95\linewidth]{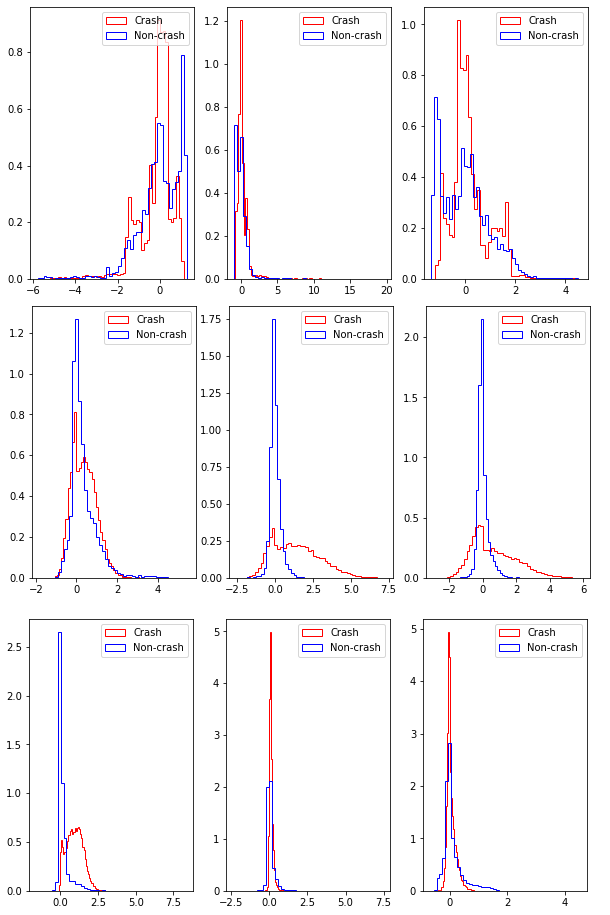}
\caption{Distributions of crash and non-crash samples for 3 random features. First row, second row, and third row relate to the features of original feature space, encoded feature space with CNN, and encoded feature space with LSTM, respectively.}
\label{fig:fspace}
\end{figure}

To determine class separability in each feature space, we measure the difference between distributions of crash and non-crash samples using well-known statistical distances such as Wasserstein metric, Bhattacharya distance, and Jensen-Shannon divergence (JS divergence).

\begin{table*}[!t]
\centering
\caption{Metrics for class separability in each feature space}
\begin{center}
\begin{tabular}{|l|c| c| c|}
\hline
  & Wasserstein distance  & Bhattacharya distance  & JS divergence  \\
  &($\times 10^{-5}$) &($\times 10^{-5}$) &($\times 10^{-5}$) \\
 \hline
Original feature space & 26 & 91 & 165\\ 
\hline
Encoded feature space with CNN  & 184 & 761 & 609\\ 
\hline
Encoded feature space with LSTM  & \textbf{409} & \textbf{1445} & \textbf{707} \\
\hline
\end{tabular}
\end{center}
\label{table:metrics}
\end{table*}

To obtain the distributions, we divide the range of values into many equal-width bins. Then we calculate the number of samples in each bin whose values are contained in that bin. Finally, the number of samples in each bin is divided by the total number of samples and the bin width so that the area under the histogram integrates to 1. In each feature space, the statistical distance between distributions of crash and non-crash samples for each feature is calculated and the average of distances is reported. The results are shown in Table~\ref{table:metrics}. Distributions of crash and non-crash samples for 3 random features of each feature space are illustrated in Figure~\ref{fig:fspace}. According to Table~\ref{table:metrics} and Figure~\ref{fig:fspace}, it is evident that distributions have greater difference in the encoded feature space with LSTM. Furthermore, since the distance between distributions has increased, it can be inferred that the correlation between features has also been reduced automatically. For example, the original feature space contained colinear features such as speed, flow, and density, whose correlations are reduced after encoding.  

According to Figure~\ref{fig:figure1}, data of 16 sensors is used in this study where each sensor provides 6 features. Furthermore, the optimal TSL value was determined to be 6 in the previous section. Therefore, each sample will have shape of (6,96) containing spatio-temporal features. This means that the dimension of original features space is 576. On the other hand, as shown in Figure~\ref{fig:arch}, the encoder outputs a feature space with dimension of 100. It can be concluded that the encoder not only increases class separability in feature space, but also reduces the dimension of data.

\subsection{Comparison with Other Artificial Intelligence (AI) Methods}
\label{SS:54}

 \begin{figure}[t]
\centering\includegraphics[width=1\linewidth]{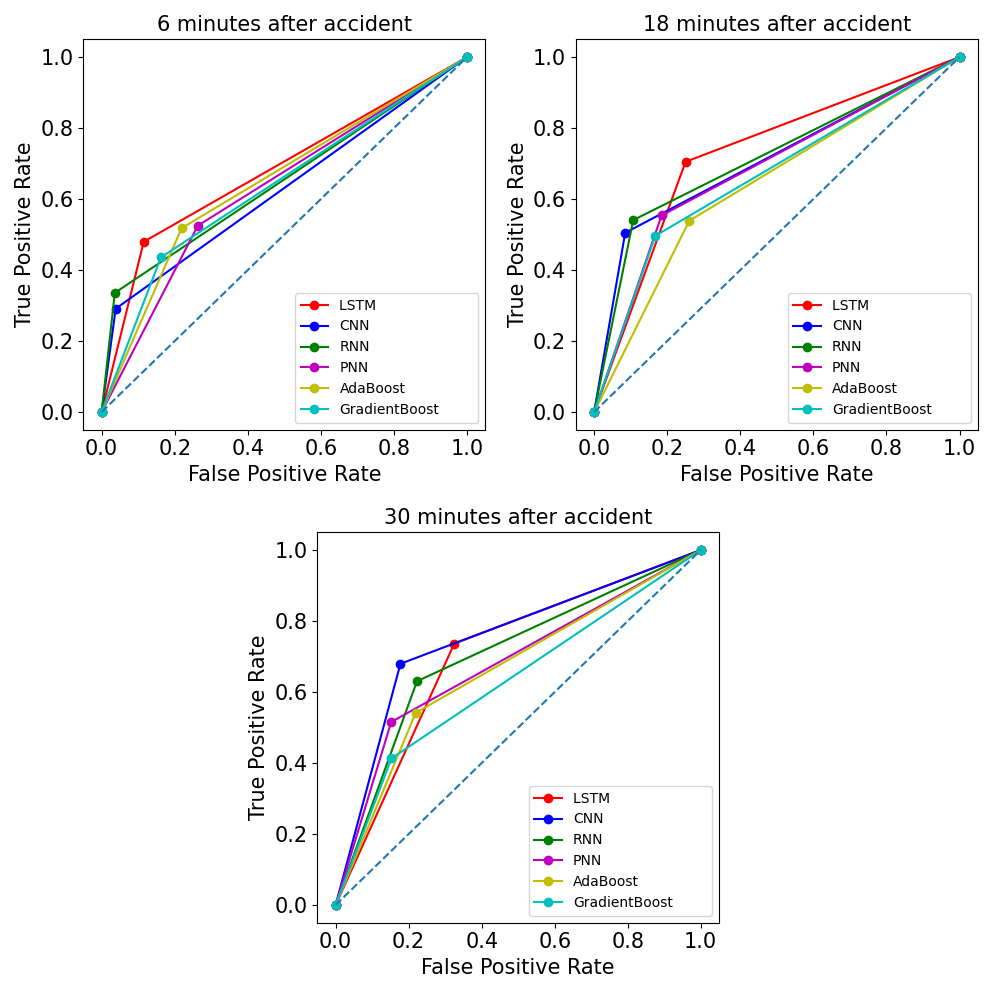}
\caption{Comparison of ROC curves for different encoders.}
\label{fig:ROC}
\end{figure}

\begin{table*}[!t]
\centering
\caption{Performance of encoders for various DAI}
\vspace{-1\baselineskip}
\begin{center}
\begin{tabular}{l l l l l l l l l l l}
\hline
 & \multicolumn{3}{c}{6 minutes after} & \multicolumn{3}{c}{18 minutes after} & \multicolumn{3}{c}{30 minutes after} \\
 \cline{2-10}
 & FPR & TPR & AUC & FPR & TPR & AUC & FPR & TPR & AUC \\
 \hline
 LSTM & 0.115 & 0.48 & \textbf{0.183} & 0.25 & \textbf{0.705} & \textbf{0.228} & 0.325 & \textbf{0.737} & 0.206\\ 
 CNN & 0.038 & 0.291 & 0.127 & \textbf{0.085} & 0.505 & 0.21 & 0.177 & 0.68 & \textbf{0.252} \\
 RNN & \textbf{0.036} & 0.336 & 0.15 & 0.107 & 0.541 & 0.217 & 0.223 & 0.63 & 0.204\\ 
 PNN & 0.263 & \textbf{0.525} & 0.130 & 0.184 & 0.554 & 0.185 & \textbf{0.152} & 0.516 & 0.182\\ 
 AdaBoost & 0.219 & 0.518 & 0.150 & 0.260 & 0.539 & 0.139 & 0.219 & 0.541 & 0.161\\
 GradientBoost & 0.162 & 0.436 & 0.137 & 0.167 & 0.496 & 0.165 & 0.153 & 0.413 & 0.130\\
\hline

\end{tabular}
\end{center}
\label{table:2}
\end{table*}

Some frameworks have reported good performance using other AI methods but on different input data. Parsa et al. \cite{parsa2019real}, for example, have reported a relatively high accuracy using PNN on weather condition, accident, and loop detector data. Therefore, in this section we compare the LSTM-based method with other AI methods that are found promising in other accident detection studies. For a fair comparison, we implemented all the models and tested them on the same site shown in Figure~\ref{fig:figure1} which is a section of a freeway with no ramps or intersections. In theory, existence of intersections or ramps will disrupt the flow and makes the scenarios more complicated. To elaborate, the out-flow of the road does not match its in-flow which causes the counts from sensor readings in different sections to be different. However, we leave these other site scenarios for our future work. For comparing the performance of the LSTM network with other deep learning methods (including CNN, PNN, RNN and ensemble methods including AdaBoost \cite{freund1997decision} and GradientBoost \cite{friedman2001greedy}), the best outcome of each model tested on a variety of sampling methods, optimizers, and network settings is reported. The ROC curves of each method for optimal TSL (TSL=6) and various DAI are shown in Figure~\ref{fig:ROC}. The Area Under Curve (AUC), False Positive Rate (FPR), and True Positive Rate (TPR) of Figure~\ref{fig:ROC} are represented in Table~\ref{table:2}. By comparing the AUC of each method, it can be concluded that LSTM is performing better at lower TTD. Furthermore, TPR of these networks at DAI=1 (6 minutes after detection) is noticeably low. Therefore, in our case study, the reasonable DAI would be 3 which means that optimal results are attained by looking into 18 minutes of loop detector data after an accident happens. It is worthwhile to mention that the results for lower TTDs are still able to detect accident patterns with reasonable accuracy. However, for comparative purposes we report 18 minutes since less false alarms and more true positives are generated at this setting.

\subsection{Generalizability of the Model}
\label{SS:55}

As concluded in the previous part, LSTM performs better at lower TTD and the optimal results were attained at DAI=3 and TSL=6. In this part, we will examine the generalizability of the LSTM encoder by training the network using optimal parameters and data of two different sites with similar characteristics. The intent is to show that a model that is trained on a site with specific characteristics in terms of road type, number of lanes, number of sensors, and distances between the sensors, can detect the accidents on other sites with similar characteristics. To this end, we study two experiments; In the first experiment, we train the network using data of site1 (shown in Figure~\ref{fig:figure1}) and test it using data of another site (site2) in which sensors are placed in a manner similar to site1. Site2 has the same characteristics as site1: Both are sections of a bidirectional freeway with limited ramps and intersection, 2 lanes, and 16 mounted loop detector sensors. The distances between the spots where sensors are mounted are not exactly the same, but yet in the same scale. In the second experiment, we train and test the network using the mix of accumulated data from both sites.

The first experiment evaluates the ability of the model to detect the accident patterns in a site different from the one it is trained on. It is trivial that the change in the distance between sensors will affect the performance of the model. However, we are interested to know how reliable the model can be when it is used to monitor a different site which has the same structure but is slightly different in arrangement of the sensors. The second experiment evaluates the model when exposed to training data that is augmented using the data of the second site described in the first experiment. Augmenting data is one way to increase the population of samples in the class with insufficient samples. However, the sensitivity of the model to the augmented data should be measured.

  \begin{figure}[t]
\centering\includegraphics[width=0.95\linewidth]{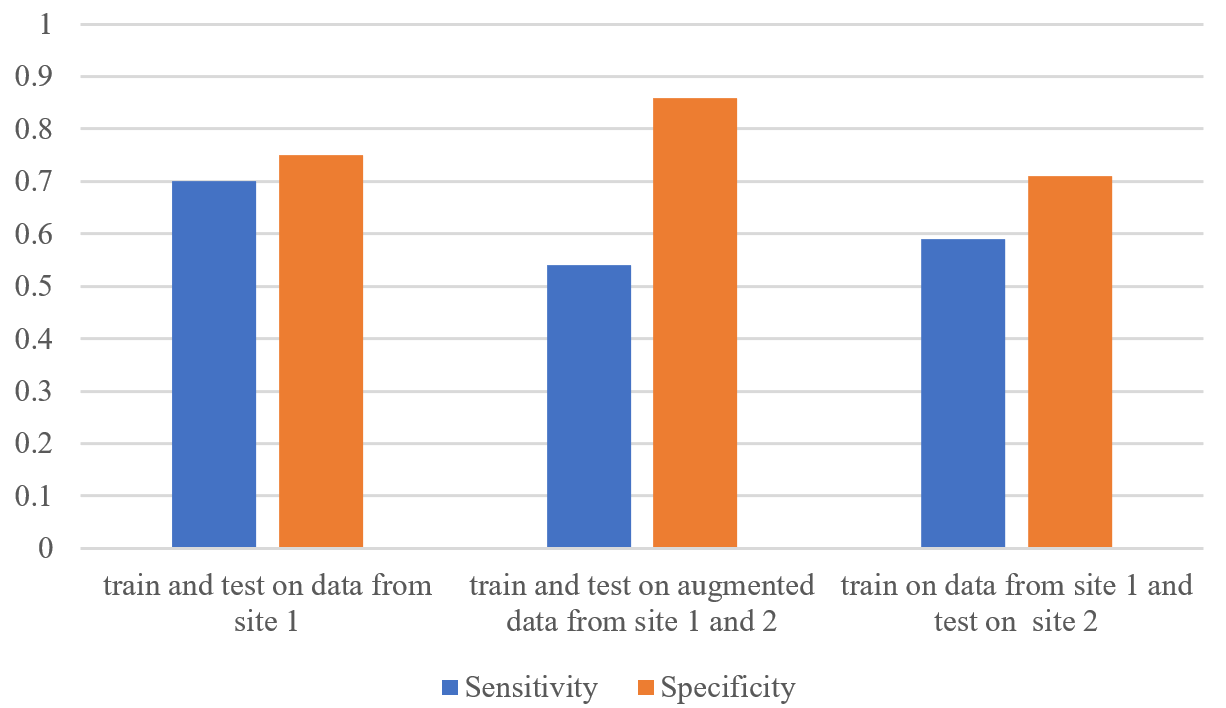}
\caption{Generalizability with augmented data and using a second site for test  }
\label{fig:gen}
\end{figure}

 Figure~\ref{fig:gen} compares sensitivity and specificity of the three experiments to test the generalizability of the model. As expected, the sensitivity of the model slightly decreases in the experiments involving the second site. The sensitivity of the model is slightly more than 0.7 before involving the second data. This value is decreased to 0.59 and 0.54 in the first and the second experiments respectively. The specificity, however, increases in the third experiment from 0.75 to 0.86, which means that we can prevent the occurrence of false alarms significantly by using augmented data.

\section{Conclusion}
\label{S:6}

Loop detector sensors, despite their noisy data and lack of high spatial coverage, are mounted on countless roads. Therefore, developing models that can yield good incident detection rate by using these sensors helps governments, to a considerable extent, obtain valuable insights until new technologies start to outnumber loop detectors on roads. Using loop detector data, we illustrated that accidents can impact the traffic flow patterns in the upstream and downstream of the accident site. Capturing these patterns in the traffic flow and associating them to the occurrence of accidents is not an easy classification problem. In this paper, we investigated an LSTM-based deep representation of imbalanced spatio-temporal traffic flow data, obtained from loop detector sensors, to study the degree to which it can increase the incident/no-incident class separability. we compared this LSTM-based deep representation with other popular AI methods like CNN, PNN, AdaBoost, and GradientBoost, and showed that the LSTM-based method increases the class separability in the encoded feature space which resulted in higher detection rate and lower false alarm rate.

A key challenge was the rareness of traffic accident events, making the collision datasets highly imbalanced. We explored different resampling techniques to identify a suitable method for dealing with heterogeneous accident data. We also explored different settings for the LSTM network as the main component of feature encoding and representation learning. The main objective was to minimize the time-to-detection of the accidents from when they happen. The results indicated that an optimal setting for time series length and data aggregation interval are 6 and 3 respectively. We also showed that the LSTM-based framework outperforms other commonly used AI methods for detecting accidents and reduces the detection time to as low as 6 minutes after crash occurrences. It is expected that this framework works better in real-time scenarios since the time stamp of pre-recorded accident data are erratic. We have shown this fact by using a tunable DoF parameter that takes the early and delayed accident flow patterns into account. Also, our experiments testing the generalizability of the model shows that the sensitivity measure decreases slightly when the model is used on a site other that the one it is trained for. Potential future works can be to further decrease the detection time, apply this method to real-time data, improve the generalizability of the model and data augmentation, test the model on parts of the freeway with ramps and intersections, and exploit other methods to map the encoded feature space to crash/non-crash outputs. 

\section*{Acknowledgement}
\label{S:Ack}

The loop detector data and accident dataset for this research is provided by Minnesota Department of Transportation, Office of Traffic Engineering.

\ifCLASSOPTIONcaptionsoff
  \newpage
\fi



\bibliographystyle{IEEEtran}
\bibliography{refs.bib}
\end{document}